%% file: main.tex
\documentclass[conference]{IEEEtran}
\IEEEoverridecommandlockouts
\usepackage{cite}
\usepackage{amsmath,amssymb,amsfonts}
\usepackage{algorithmic}
\usepackage{graphicx}
\usepackage{textcomp}
\usepackage{xcolor}
\def\BibTeX{{\rm B\kern-.05em{\sc i\kern-.025em b}\kern-.08em
    T\kern-.1667em\lower.7ex\hbox{E}\kern-.125emX}}
\usepackage{stfloats}
\usepackage{url}
\usepackage{verbatim}
\usepackage{cite}
\usepackage{booktabs}
\usepackage{amsfonts}
\usepackage{nicefrac}
\usepackage{longtable}
\usepackage{pifont}
\usepackage{circledsteps}
\usepackage{xspace}
\usepackage{tabularray}
\usepackage{svg}
\newcommand{\sysname}[0]{{\sc DGRAG}\xspace}
    
\begin{document}

\title{DGRAG: Distributed Graph-based Retrieval-Augmented Generation in Edge-Cloud Systems}

\author{\IEEEauthorblockN{Wenqing Zhou, Yuxuan Yan, Qianqian Yang}
\IEEEauthorblockA{\textit{Zhejiang University}}
}

\maketitle

\begin{abstract}
Retrieval-Augmented Generation (RAG) improves factuality by grounding LLMs in external knowledge, yet conventional centralized RAG requires aggregating distributed data, raising privacy risks and incurring high retrieval latency and cost. We present DGRAG, a distributed graph-driven RAG framework for edge–cloud collaborative systems. Each edge device organizes local documents into a knowledge graph and periodically uploads subgraph-level summaries to the cloud for lightweight global indexing without exposing raw data. At inference time, queries are first answered on the edge; a gate mechanism assesses the confidence and consistency of multiple local generations to decide whether to return a local answer or escalate the query. For escalated queries, the cloud performs summary-based matching to identify relevant edges, retrieves supporting evidence from them, and generates the final response with a cloud LLM. Experiments on distributed question answering show that DGRAG consistently outperforms decentralized baselines while substantially reducing cloud overhead.
\end{abstract}


\input{sections/Introduction}
\input{sections/RelatedWorks}
\input{sections/SystemDesign}
\input{sections/Evaluation}
\input{sections/Conclusion}

\bibliographystyle{IEEEtran}
\bibliography{ref}

\end{document}

%% file: sections/Introduction.tex
\section{Introduction}
\begin{figure*}[htp]
    \centering
    \includegraphics[width=0.95\linewidth]{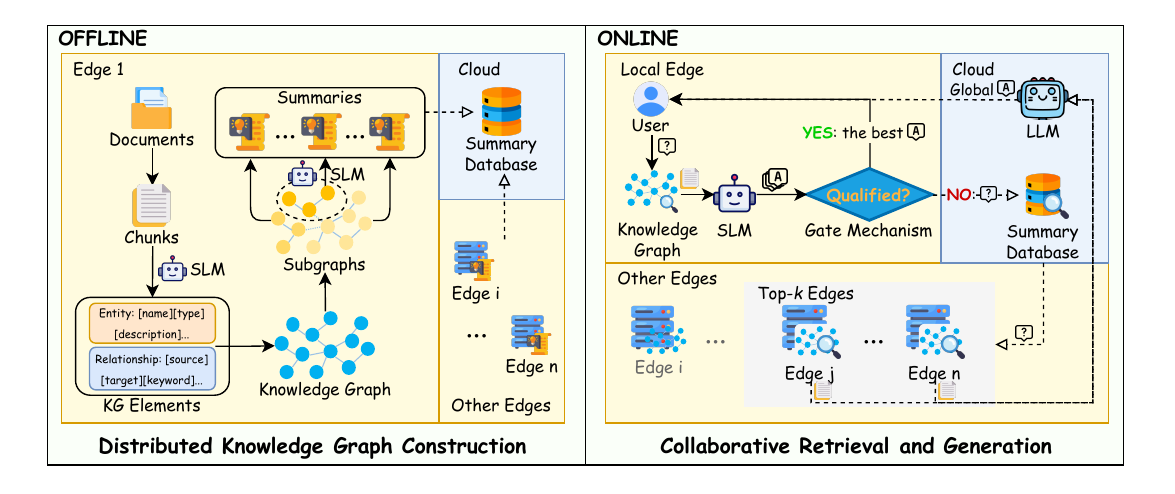}
    \caption{Overview of the DGRAG framework, which includes two main phases: Distributed Knowledge Graph Construction (left) and Collaborative Retrieval and Generation (right).}
    \label{fig:overview}
\end{figure*}

Large language models (LLMs) have achieved remarkable success in natural language understanding and generation, owing to extensive pre-training on massive corpora\cite{gao2021condenser} and advanced fine‑tuning techniques\cite{zhou2024comprehensive}. Despite this fluency and broad applicability, the knowledge within LLMs is inherently constrained by the scope and recency of their training data, which contributes to the generation of fluent yet factually incorrect text—a phenomenon often called ``hallucination'' \cite{huang2023survey,xu2024hallucination}. Moreover, LLMs frequently fall short when addressing highly specialized domains or rapidly changing information contexts. To overcome these limitations, Retrieval-Augmented Generation (RAG)\cite{lewis2020retrieval} has emerged as a compelling architectural paradigm. By coupling the generative capability of LLMs with on-demand retrieval from external knowledge sources, RAG enhances factual fidelity and domain adaptability of generated outputs.

While RAG significantly enhances language model factuality, most existing systems rely on a centralized architecture, where all external knowledge is stored in a unified repository. This ``all-in-one'' approach leads to high communication and computational overhead, as well as privacy and compliance risks~\cite{gao2023retrieval,hofstatter2023fid}, especially in sectors like healthcare, finance and public administration. Furthermore, centralized systems struggle with scalability and real-time responsiveness~\cite{zhu2024accelerating}, making them unsuitable for large-scale, dynamic data environments. Conversely, fully decentralizing RAG to allow each edge node to independently perform retrieval and generation addresses privacy concerns but results in limited knowledge coverage and reduced generation quality, especially for complex or cross-domain queries. Thus, future RAG frameworks need to balance centralization and decentralization: preserving distributed data storage to protect privacy without uploading raw data, while enabling cross-node collaboration to overcome edge limitations.

Guided by this principle, we propose DGRAG, a distributed, graph-based RAG system designed for privacy-preserving, knowledge-collaborative generation in edge–cloud environments. Each edge node maintains a local knowledge base organized as a knowledge graph (KG), enabling efficient local retrieval. The cloud server coordinates these edge KGs via shared subgraph-level summaries, which are uploaded periodically instead of raw data to support privacy and global indexing. During inference, a gate mechanism decides if a query can be answered locally based on similarity and confidence of multiple local responses. Queries beyond the scope of local knowledge are selectively escalated to the cloud. There, a cross-edge retrieval mechanism identifies relevant edges, aggregates evidence, and guides the cloud LLM to generate a comprehensive response. This approach balances local autonomy with global collaboration, ensuring privacy, efficiency, and high-quality retrieval-augmented generation.

The main contributions of this paper are summarized as follows: 
\begin{itemize}
    \item [1)] We introduce subgraph summaries to facilitate efficient information sharing among distributed knowledge graphs. By organizing local knowledge into knowledge graphs and sharing these summaries, \sysname reduces communication overhead between edge devices and the cloud, alleviating memory burdens on centralized storage while ensuring data privacy.
    \item [2)] We design an edge-to-cloud workflow with a gate mechanism that balances efficiency and quality in the RAG process. This workflow leverages the real-time processing power of edge devices and the cloud’s computational resources for more complex queries. The gate mechanism dynamically evaluates local model performance and knowledge adequacy, adjusting task allocation between the edge and cloud based on batch-generated response similarity.
    \item [3)] We develop a cross-edge retrieval mechanism that enables knowledge collaboration across multiple edge nodes to enhance cloud-based generation. When an edge node faces resource limitations, the cloud can augment inference using more powerful models and retrieve knowledge from other edges, ensuring both task efficiency and the generation of more accurate, comprehensive responses.
\end{itemize}

%% file: sections/RelatedWorks.tex
\section{Related Work}

\subsection{RAG in distributed environments}
While significant advancements have been made in improving retrieval efficiency and generation quality within centralized RAG systems, there has been limited exploration of RAG application in decentralized, edge-based environments.

In distributed knowledge environments, effective response generation depends critically on mechanisms for filtering knowledge sources and enabling cross-node collaboration. Several works explore RAG designs that support cooperation among edge nodes. DRAG~\cite{xu2025distributed} proposes the Topic-Aware Random Walk (TARW), which discovers relevant peers and retrieves knowledge in a peer-to-peer network without centralized storage. CoEdge-RAG~\cite{hong2025coedge} introduces a semantics-aware hierarchical scheduler that identifies query intent on the fly and routes queries to appropriate edge nodes for balanced and efficient processing. The authors of \cite{lu2025decentralized} present a decentralized RAG framework that assesses sentence-level importance and updates source reliability through a feedback-driven scoring scheme secured by blockchain smart contracts. Other approaches utilize a central coordinator or cloud server to enable collaborative or hierarchical retrieval. EACO-RAG~\cite{li2024eaco} employs a contextual multi-armed bandit with Safe Online Bayesian Optimization to jointly optimize retrieval and generation, improving latency, cost efficiency, and accuracy. DRAGON~\cite{liu2025efficient} leverages cloud computing and parallel token generation, supported by a Speculative Aggregation algorithm, to enhance the performance of small language models on edge devices.

Regarding data privacy and security, existing research has primarily focused on securing data transmission and access processes. C-FedRAG~\cite{addison2024c} designs a confidential computing framework that uses federated mechanisms for retrieval and generation across multiple data providers, ensuring privacy while enabling LLM-based knowledge access. The authors of~\cite{11014986} introduce the External Retrieval Interface (ERI), a standardized API that allows data providers to offer retrieval functions without exposing raw data, while enabling model providers to implement custom access control policies.

Most edge-based distributed RAG systems rely on vector retrieval, but this approach overlooks semantic relationships and struggles with cross-device knowledge linking, particularly as knowledge bases expand. Moreover, direct text data transfers also lead to significant communication overheads, limiting practicality in real-world applications. These challenges motivate our work to adopt knowledge graphs as a structured foundation to achieve more efficient, expressive, and scalable retrieval in distributed environments.

\subsection{Graph-based RAG}
In recent years, RAGs have become a key technique for enhancing LLMs. Traditional RAG systems rely on dense retrieval methods based on vector similarity, which often struggle to capture complex relationships and deep semantic connections. To address these limitations, graph-based RAG has emerged, leveraging the powerful representational capabilities of knowledge graphs (KGs) to enable more precise and contextually rich retrieval.

GraphRAG~\cite{edge2024local} pioneered graph-based RAG by integrating KGs into the retrieval process, enhancing the diversity and coverage of retrieved documents while reducing computational cost. This paradigm has since been streamlined by LightRAG~\cite{guo2024lightrag} and MiniRAG~\cite{fan2025minirag}. LightRAG~\cite{guo2024lightrag} introduces a dual-level retrieval framework capturing both entity-level and thematic information, whereas MiniRAG~\cite{fan2025minirag} designs a semantic-aware heterogeneous graph index and topology-enhanced retrieval tailored for small language models (SLMs) in resource-limited settings. Building on these developments, the authors of~\cite{zhao20252graphrag} propose a lightweight graph-based index with an adaptive retrieval mechanism that dynamically switches between local and global search based on entity connectivity, enabling faster indexing and more efficient evidence selection while preserving answer quality.

A number of studies optimize graph-based RAG by combining graph signals with textual information to strengthen evidence discovery.  GRAG~\cite{hu2024grag} enhances the capability of LLMs to process graph-structured data by incorporating both textual and topological information at a finer granularity. HybridRAG~\cite{sarmah2024hybridrag} merges vector-based and graph-based RAG to strike a balance between efficiency and accuracy, improving knowledge retrieval in complex documents. Other works focus on improving graph structures and corresponding retrieval mechanisms. PathRAG~\cite{chen2025pathrag} introduces a flow-based pruning algorithm and path-based prompting to guide LLMs, reducing redundancy and generating more coherent responses. NodeRAG~\cite{xu2025noderag} develops a graph-centric RAG framework based on a carefully designed heterogeneous graph that unifies diverse information into a fully nodalized structure, enabling fine-grained, explainable, and hierarchically coherent retrieval. Similarly, ClueRAG~\cite{su2025clue} constructs a multi-partite graph linking chunks, knowledge units, and entities, and retrieves evidence through a query-driven iterative process (Q-Iter), anchoring on entities and performing constrained graph traversal. Beyond text-only settings, recent studies extend graph-based RAG to richer multimodal scenarios. AVA ~\cite{yanava} constructs Event Knowledge Graphs from video streams, coupling graph-based retrieval with agentic link traversal to enable multi-hop temporal reasoning and query-focused summarization. Vgent ~\cite{shen2025vgent} represents long videos as structured graphs of clip-level visual entities and relations, introducing a graph-based retrieval–reasoning pipeline with structured query verification to refine retrieved clips and aggregate multimodal evidence.

In distributed edge-cloud environments, knowledge graphs are ideal for efficiently representing and sharing data. It has proven to be highly effective in capturing the core entities and relationships within data\cite{peng2024graph, han2024retrieval}, a feature that can be effectively leveraged in collaborative RAG systems. They enable high-level semantic abstraction of local data, facilitating efficient information exchange and joint reasoning between the cloud and edge nodes. This approach reduces communication overhead, preserves privacy, and enhances system scalability.

%% file: sections/SystemDesign.tex
\section{System design}

\subsection{Overview}
We consider a distributed system comprising $N$ edge devices, each deployed with a small language model (SLM) based on its resource constraints, and a cloud server, as shown in Fig.~\ref{fig:overview}, where the proposed DGRAG enables collaborative, knowledge-enhanced RAG across edges. DGRAG operates in two phases: the \textbf{Distributed Knowledge Graph Construction} phase and the \textbf{Collaborative Retrieval and Generation} phase. In the Distributed Knowledge Graph Construction phase, each edge node organizes its local knowledge to a knowledge graph and sends summaries of the subgraphs to the server. During the Collaborative Retrieval and Generation phase, once an edge user sends a query, the local SLM first tries to answer it; a gate mechanism then determines whether the generated response is satisfactory. If not, the query is forwarded to the cloud server. The server then attempts to identify other edge nodes that may possess relevant knowledge by consulting the summary database. Using the retrieved knowledge from the most related edge nodes, the server answers the query with a global large language model and sends the response back to the edge. We will detail each step of these two phases in the following.

\subsection{Distributed Knowledge Graph Construction}
The \textbf{Distributed Knowledge Graph Construction} phase consists of the following three steps.

\textbf{\ding{172} Edge Graph Knowledge Extraction.} In this step, each edge node first performs document segmentation and linguistic analysis on its local dataset. Specifically, each original document is split into text chunks. The edge SLM extracts entities (e.g., persons, places, events, objects, etc.) and the relationships between them from the text chunks, which together constitute the basic units of edge knowledge graphs (Edge KGs), with entities as nodes and relationships as edges. After this process, the vector embeddings of all entities, relationships, along with their attributes, and text chunks, are stored in three separate vector databases. In addition, a graph database is generated to capture the Edge KG. Together, these three vector databases and the graph database form a complete edge knowledge base at each edge node, which can be later used for RAG.

\textbf{\ding{173} Topology-based Graph Partitioning.}
\sysname employs the Leiden algorithm \cite{traag2019louvain} to iteratively partition the KG, constructed in the previous step, at each edge node into subgraphs that optimize modularity, ensuring that entities within the same subgraph are more densely connected than those across different subgraphs. Neighboring subgraphs are then merged if they are too small to meet a predefined size threshold. Next, the entities (i.e., nodes) and relationships (i.e., edges) in each subgraph, along with their attributes, are transformed into plain text, which is then input into the local SLM to generate a summary of the subgraph. These summaries are subsequently sent to the server for aggregation. Notably, subgraph summaries concisely capture the core content of each subgraph—including subject areas, main entities, and key relationships—without disclosing too many details, facilitating information sharing in a distributed environment.

\textbf{\ding{174} Federated Knowledge Aggregation.} Each edge then sends its subgraph summaries to the server, where all subgraph summaries from the edge nodes are stored in a global summary vector database to form a comprehensive overview of the knowledge across the entire edge network.

\begin{figure*}[htp]
    \centering
    \includegraphics[width=0.9\linewidth]{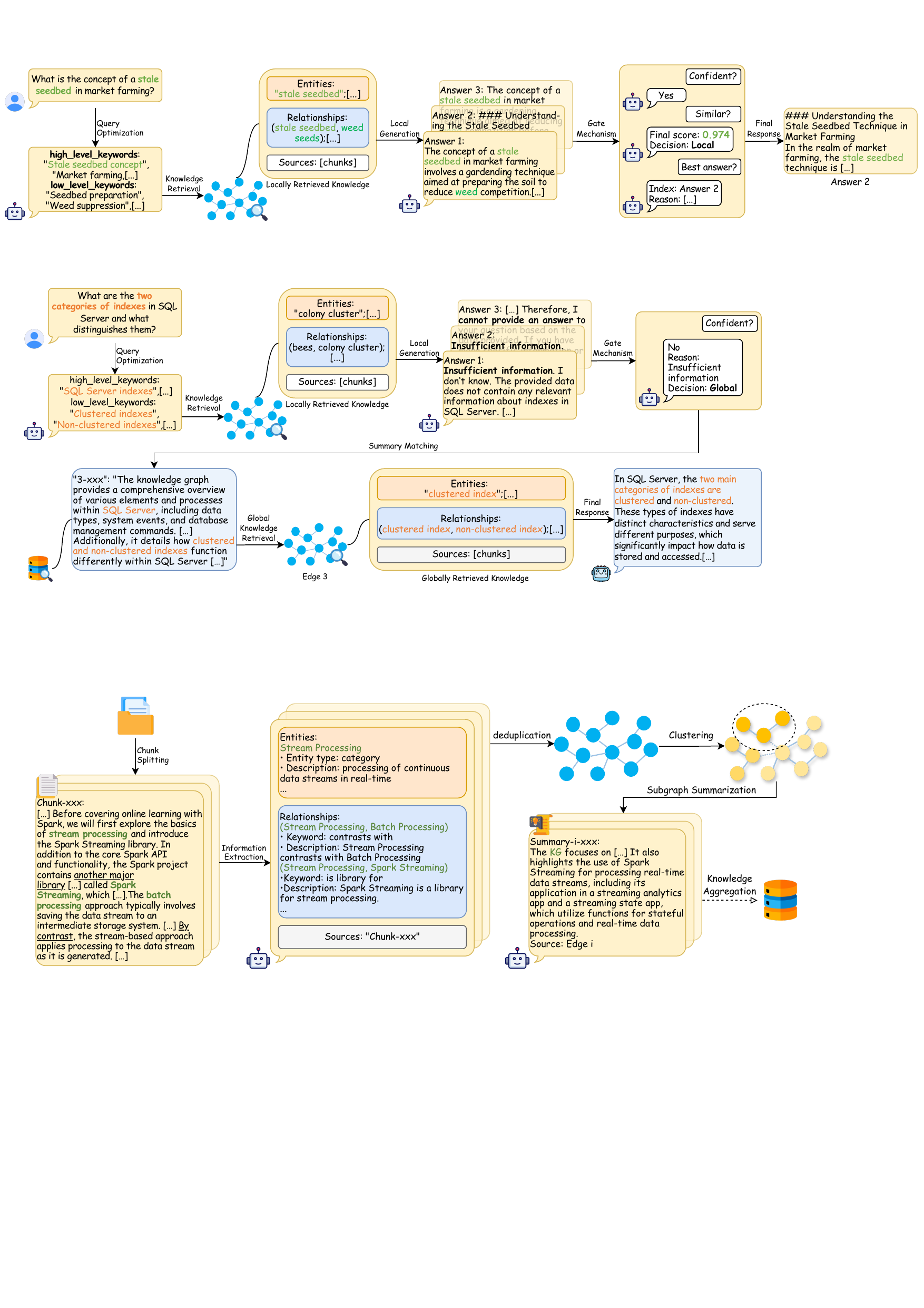}
    \caption{A real case of the Distributed Knowledge Graph Construction pipeline from the computer science domain.}
    \label{fig:subsummary}
\end{figure*}

Fig. \ref{fig:subsummary} provides a detailed example of how a subgraph summary is generated, from the construction and partitioning of an edge KG to the summarization process. Subgraph summaries generated at each edge device are ultimately centralized and stored in the cloud summary vector database.

\subsection{Collaborative Retrieval and Generation}
After the knowledge graph construction described above, when an edge user initiates a query, DGRAG performs \textbf{Collaborative Retrieval and Generation} in the following three steps.

\textbf{\ding{175} Local Query.} The edge user who initiates a query first attempts to generate a response using the local SLM, which adopts a dual-level retrieval mechanism \cite{guo2024lightrag}. First, the system matches the most similar entities and relationships to the query from the local vector database and then expands the context by retrieving their neighbors from the graph database to enrich the knowledge. Additionally, the original text chunks associated with the retrieved entities and relationships are also fetched from the database. All of this retrieved information, along with the original query, is fed into the local SLM to generate a preliminary local response, which serves as a basis for further refinement. Note that the query may fall outside the domain of the local knowledge base, potentially leading to incorrect responses from the local SLM. To enable the proposed gating mechanism, which will be introduced next, to assess whether a response is satisfactory, \sysname adopts batch inference to generate multiple candidate responses at once, rather than producing a single response.

\textbf{\ding{176} Gate Mechanism.} The gate mechanism at the edge node evaluates whether the query exceeds the scope of local knowledge or processing capabilities, thus determining whether the global cross-edge retrieval is necessary. This evaluation involves three steps: 1) \textbf{Confidence Detection}: We first detect whether local responses express a lack of confidence or information—such as phrases like `insufficient information' or `need more details'—using the local SLM. If the responses exhibit a certain level of uncertainty, the query is sent to the cloud for global retrieval and generation; otherwise, it proceeds to the next step; 2) \textbf{Similarity Evaluation}: Similarity Evaluation involves a detailed assessment of the generated local responses to determine their consistency. The SLM evaluates the multi-dimensional similarity of the responses on diverse metrics, including cosine similarity, the Jaccard index, and semantic consistency of core claims. Cosine similarity measures the similarity between the vector representations of the responses, capturing their semantic overlap. The Jaccard index evaluates the similarity based on the intersection over union of the sets of words or phrases in the responses as a reference at the lexical level. Semantic consistency checks the alignment of core claims and assertions across the responses, ensuring that they convey a coherent and consistent message. It finally generates an average of these sub-dimensional scores to produce an overall similarity score. 3) \textbf{Similarity-based Selection}: If the overall similarity score exceeds a predefined threshold, the locally generated responses are deemed satisfactory. The SLM then selects the best response to answer the query. Otherwise, the original query is sent to the server for further retrieval and generation. This mechanism is motivated by the observation that, in the absence of sufficient knowledge, generated responses tend to exhibit a high degree of hallucination, resulting in low similarity across multiple outputs. In contrast, correct responses across multiple inferences are typically highly similar.

\begin{figure*}[htp]
    \centering
    \includegraphics[width=0.94\linewidth]{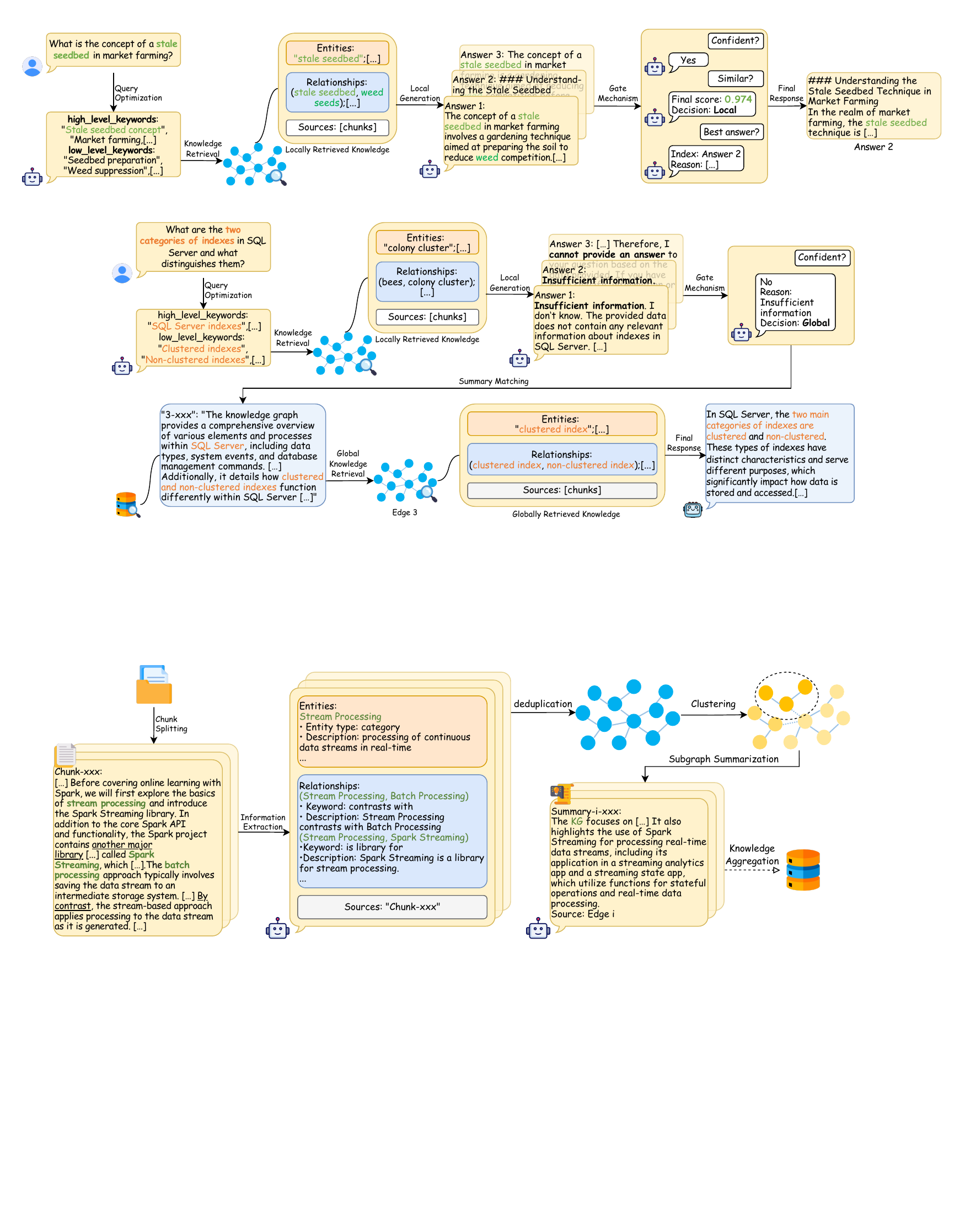}
    \caption{A real case of \sysname workflow in adequate local knowledge. Notably the query in the illustrated case originates from the agricultural domain, and the querying edge node also possesses knowledge of agricultural domain.} 
    \label{fig:collaborative rag1}
\end{figure*}

\begin{figure*}[htp!]
    \centering
    \centerline{\includegraphics[width=1.12\textwidth]{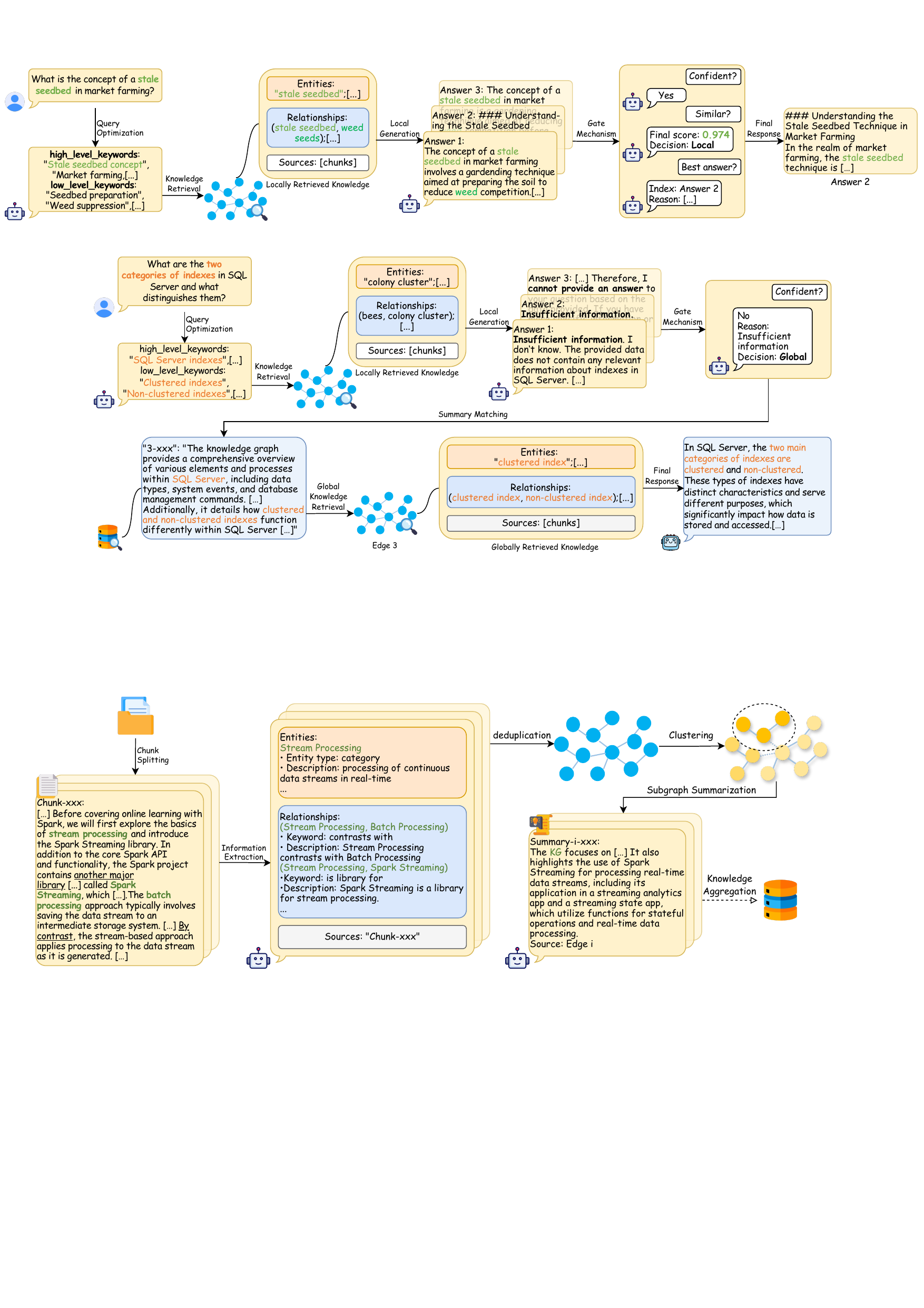}}
    \caption{A real case of \sysname workflow under insufficient local knowledge. Notably the query in the illustrated case originates from the computer science domain, and the querying edge node solely possesses knowledge of agricultural domain.}
    \label{fig:collaborative rag2}
\end{figure*}

\textbf{\ding{177} Cross-edge Retrieval Mechanism.} Once the cloud server receives a query from an edge node, it initiates cross-edge collaborative retrieval and generation. First, the cloud queries the global summary database to identify the top-$m$ most relevant summaries related to the query, and then sends retrieval requests—along with the original query—to the top-$k$ edge nodes that own the corresponding summaries. Second, the edge nodes receiving the requests perform knowledge retrieval as they would for a local query and return highly relevant knowledge back to the cloud. Finally, the LLM at the cloud server generates a global response using the aggregated knowledge and the original query. The server then sends the final response back to the originating edge user.

Fig.~\ref{fig:collaborative rag1} and Fig.~\ref{fig:collaborative rag2} outline the specific workflow of \sysname for completing the RAG process in different cases of local knowledge. Fig.~\ref{fig:collaborative rag1} illustrates the workflow of \sysname when sending an agriculture-related query at the agricultural edge node which locally possesses relevant knowledge. In adequate local knowledge, The whole workflow processes only at the querying edge. After the Local Query step, the gate mechanism determines that local responses have a high degree of similarity, whereby local knowledge is considered sufficient and the best local answer is selected as the final response. Fig.~\ref{fig:collaborative rag2} shows a case where an agricultural edge node sends a computer science-related query. Obviously, the querying edge node lacks relevant knowledge in the subject domain. Under such conditions, the local response exhibits lack of confidence. The gate mechanism interprets this as an indication of inadequate local knowledge and decides to transfer the query to the cloud server. The cloud server augments response generation through the cross-edge collaborative retrieval mechanism. After retrieving knowledge from the most relevant edge to the query, the cloud LLM generates the final response based on the aggregated knowledge context.

%% file: sections/Evaluation.tex
\section{Evaluation}
\label{sec:eval}

\subsection{Implementation}
We leverage Qwen2.5-14B \cite{bai2023qwen} as the edge-side SLM, using its AWQ-quantized variant to reduce memory footprint and enable efficient deployment on resource-constrained edge devices. For cloud-side generation, we employ Qwen-Max \cite{bai2023qwen} as the LLM. We adopt Qwen3-Embedding-0.6B \cite{bai2023qwen} as the embedding model, and use the vLLM framework \cite{kwon2023efficient} to support batch inference and accelerate both local and cloud-side generation. For data storage, we use NetworkX \cite{SciPyProceedings_11} to maintain the graph databases and Nano-vectordb to store vector embeddings.

\subsection{Settings}

\textbf{Datasets.}
To simulate heterogeneous yet potentially related knowledge distributed across edge nodes, we adopt the UltraDomain benchmark \cite{qian2024memorag} as our dataset. UltraDomain is a comprehensive benchmark designed for long-context, multi-domain Q\&A and RAG evaluation, with queries requiring global comprehension of extended contexts and multi-hop reasoning. It contains 18 datasets curated from 428 college-level textbooks, with individual contexts reaching up to one million tokens.

\textbf{Metrics.}
We adopt two complementary evaluation metric types: relative evaluation metrics with a LLM evaluator and reference-based evaluation metrics compared with the ground-truth references. For the relative evaluation, we employ the evaluation metrics used in LightRAG\cite{guo2024lightrag} and use Qwen-Max\cite{bai2023qwen} as the evaluator. The primary metrics include: (1) \emph{Comprehensiveness}, measuring the degree to which a response covers all relevant aspects of a query; (2) \emph{Diversity}, evaluating the breadth of perspectives and insights expressed in the response; (3) \emph{Empowerment}, assessing how effectively the response helps the reader understand the question and its underlying concepts; and (4) \emph{Overall}, reflecting combined performance across all three dimensions. In addition to the relative evaluation, we also incorporate BERTScore Recall \cite{zhang2019bertscore} and Rouge-L Recall \cite{lin2004rouge} as reference-based evaluation metrics to assess similarity to the ground-truth references. Reporting both metrics allows us to capture the intrinsic performance of \sysname and its improvements over the baselines.

\begin{table*}[ht]
\centering
\caption{Average win rates of \sysname compared to baseline RAG approaches in domain-specific Q$\&$A tasks.}
\label{tab:t1}
\begin{tabular}{lcccccccc}
\toprule[1.5pt]
\multicolumn{1}{l}{\textbf{Scenario}}                                                    & \multicolumn{4}{c}{\textbf{Within-domain}}                         & \multicolumn{4}{c}{\textbf{Out-of-domain}}\\ \midrule
\multicolumn{1}{l}{\textbf{\begin{tabular}[l]{@{}l@{}}Method\\ Comparison\end{tabular}}} & \begin{tabular}[c]{@{}c@{}}DGRAG vs. \\ Naïve RAG\end{tabular} & \begin{tabular}[c]{@{}c@{}}DGRAG vs.\\ Local RAG\end{tabular} & \begin{tabular}[c]{@{}c@{}}DGRAG vs.\\ Cloud RAG\end{tabular} & \begin{tabular}[c]{@{}c@{}}DGRAG vs.\\ Cent. RAG\end{tabular} & \begin{tabular}[c]{@{}c@{}}DGRAG vs. \\ Naïve RAG\end{tabular} & \begin{tabular}[c]{@{}c@{}}DGRAG vs. \\ Local RAG\end{tabular} & \begin{tabular}[c]{@{}c@{}}DGRAG vs.\\ Cloud RAG\end{tabular} & \begin{tabular}[c]{@{}c@{}}DGRAG vs.\\ Cent. RAG\end{tabular} \\ \midrule
Comprehensiveness                                                                        & 68.8\%                                                         & 80.0\%                                                        & 45.3\%                                                        & 76.5\%                                                        & 85.5\%                                                         & 91.0\%                                                         & 61.8\%                                                        & 76.3\%                                                        \\
Diversity                                                                                & 48.5\%                                                         & 66.0\%                                                        & 28.8\%                                                        & 49.3\%                                                        & 82.8\%                                                         & 90.5\%                                                         & 45.5\%                                                        & 67.5\%                                                        \\
Empowerment                                                                              & 58.0\%                                                         & 72.5\%                                                        & 35.3\%                                                        & 65.8\%                                                        & 82.8\%                                                         & 91.0\%                                                         & 58.8\%                                                        & 73.3\%                                                        \\ \midrule
\textbf{Overall}                                                                         & \textbf{59.3\%}                                                & \textbf{73.3\%}                                               & \textbf{36.3\%}                                               & \textbf{66.8\%}                                               & \textbf{83.3\%}                                                & \textbf{91.0\%}                                                & \textbf{59.5\%}                                               & \textbf{73.5\%}                                               \\ \bottomrule[1.5pt]
\end{tabular}
\end{table*}

\textbf{Baselines.}
We compare \sysname against four RAG approaches in distributed Q$\&$A settings. (1) \emph{Naïve RAG} serves as the standard baseline, where raw texts are chunked and stored in a vector database, and retrieval is performed by embedding the query and selecting the most similar chunks for generation.  (2) \emph{Local RAG} corresponds to a standalone implementation of the local query process in \sysname. It builds a local knowledge base following the procedure in \textbf{Edge Graph Knowledge Extraction}, retrieves relevant knowledge as described in \textbf{Local Query}, and generates responses solely with the edge SLM. (3) \emph{Cloud RAG} represents a standalone implementation of the global query process, where queries directly trigger the Cross-edge Retrieval Mechanism and responses are generated by the cloud LLM. (4) \emph{Centralized RAG (Cent. RAG)} aggregates all edge knowledge in the cloud to form a centralized knowledge base, applies the same retrieval pipeline as in Local Query, and uses the cloud LLM for final generation. (5) \emph{\sysname} denotes our proposed distributed approach.

\textbf{Evaluation scenarios.} To assess DGRAG under varying scales of edge-cloud systems and retrieval sources, we designed two evaluation scenarios. (1) \textbf{Domain-specific Q$\&$A}: This task directly evaluates DGRAG's performance by limiting cross-edge retrieval to a single edge node. We select the agriculture, art, cooking, and computer science (CS) datasets of UltraDomain. In this setup, the edge-cloud system consists of four edge nodes, each holding a complete domain context and operating as a standalone knowledge base. The number of entities in a single subgraph is set as 40. (2) \textbf{Mixed-domain Q$\&$A}: This evaluation scales up the system and the complexity of knowledge sources. We distribute the 10 smallest UltraDomain datasets across ten edges and the domain contexts follow a Dirichlet distribution with parameter $\alpha$ = 10. Due to the larger knowledge base, the top-$m$ value for Summary Matching is increased to 5. Additionally, thresholds are introduced to control retrieval volume in both the \textbf{Summary Matching} and \textbf{Global Retrieval} sessions, where only summary or knowledge surpassing the threshold is considered. Both thresholds are set to 0.4.

\subsection{Overall performance}

\begin{figure}[htp]
    \centering
    \includegraphics[width=1\columnwidth]{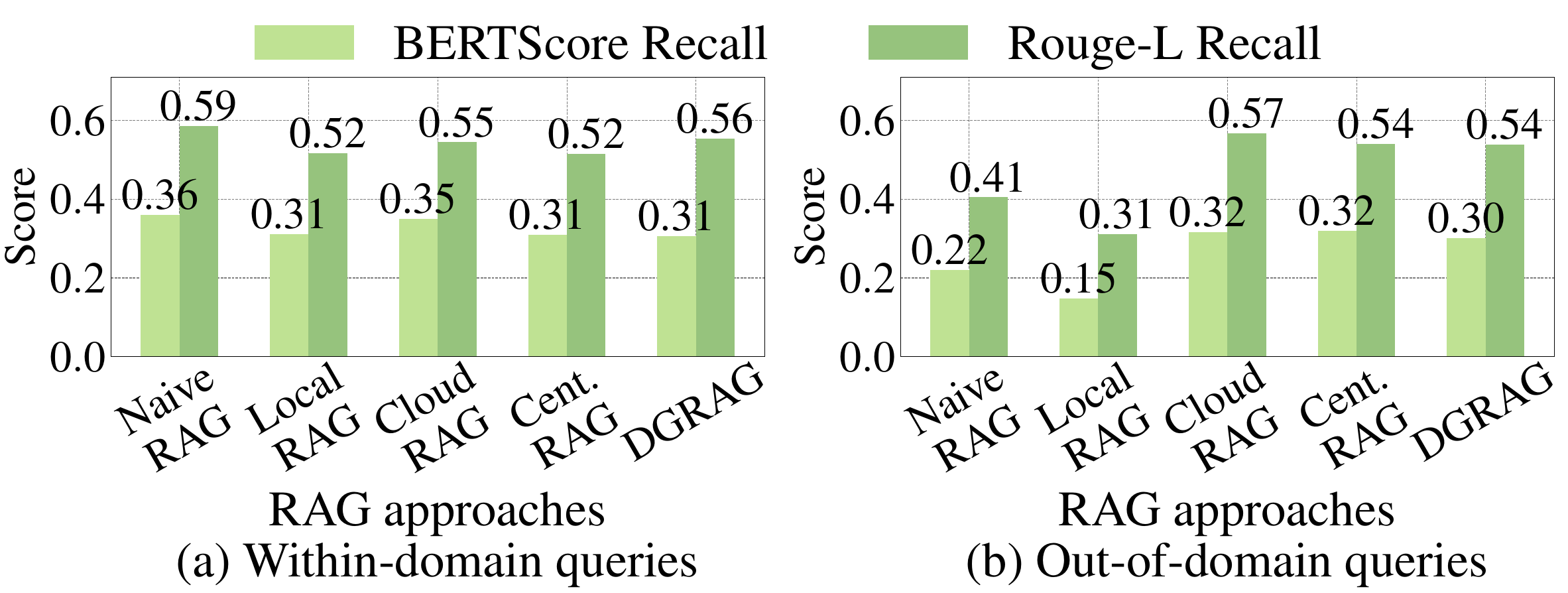}
    \caption{BERTScore Recall and Rouge-L Recall of responses by different RAG approaches in domain-specific Q$\&$A tasks: (a) within-domain queries; (b) out-of-domain queries.}
    \label{fig:recall}
\end{figure}

\begin{figure}[htp]
    \centering
    \includegraphics[width=0.85\columnwidth]{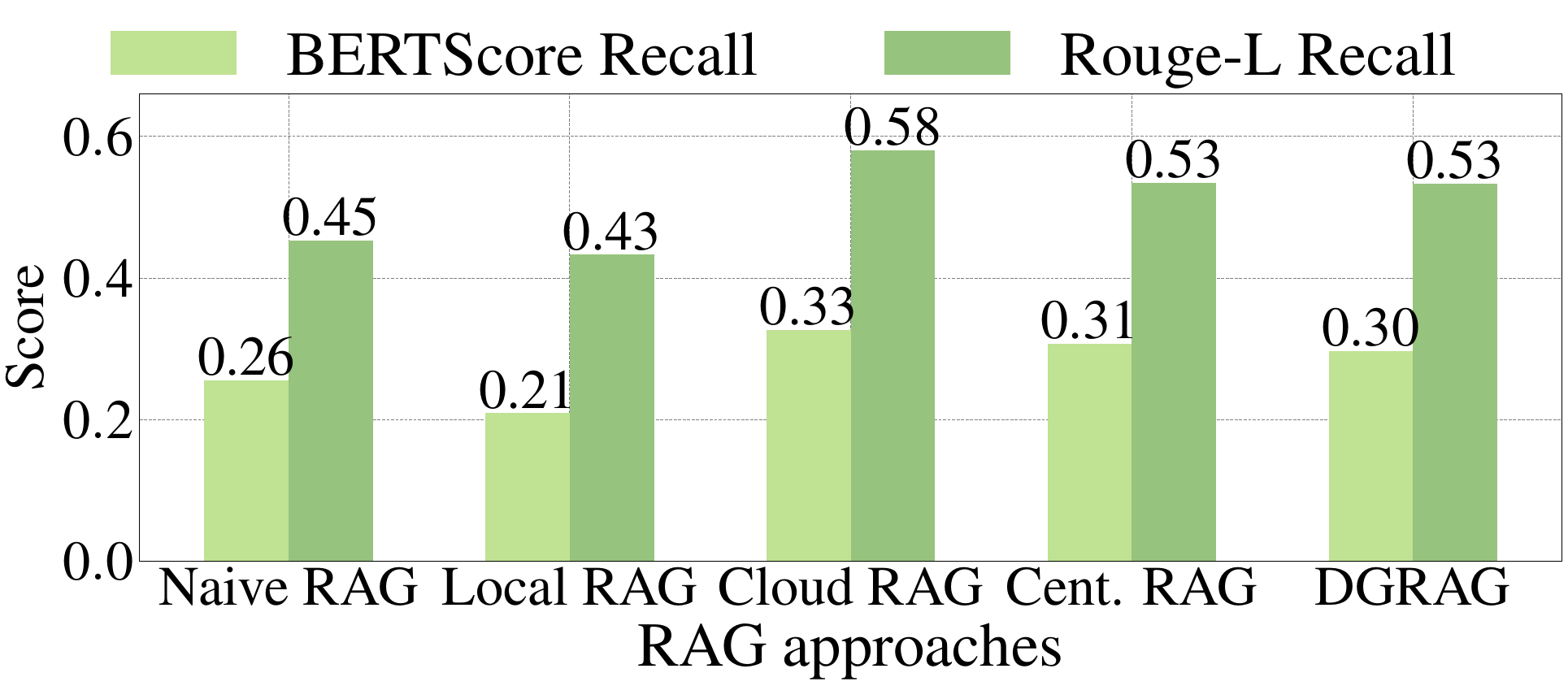}
    \caption{BERTScore Recall and Rouge-L Recall of responses by different RAG approaches in mixed-domain Q$\&$A tasks.}
    \label{fig:main_thd_exp2}
\end{figure}

We design query settings tailored to different knowledge-distribution scenarios. For the domain-specific Q$\&$A, where each edge node holds domain-specialized knowledge, we select 100 queries per domain for a total of 400 queries and divide the queries assigned to each edge node into \textit{in-domain} and \textit{out-of-domain} scenarios. In-domain queries originate from the edge node’s own knowledge domain, while out-of-domain queries come from the remaining three domains. Each of the four edge nodes is required to answer both 100 in-domain queries and 100 out-of-domain queries. This setup enables an explicit evaluation of the sensitivity of the gate mechanism and the accuracy of Summary Matching session. For the mixed-domain Q$\&$A, edge knowledge is heterogeneous. Each edge node is required to answer 100 queries drawn exclusively from one of the ten domains, a setup that contrasts with the domain-specific setting and better reflects real-world data distributions. We report DGRAG’s average win rates over baseline RAG approaches and the recall metrics of all generated answers against ground truth, as shown in Table~\ref{tab:t1}, Fig.~\ref{fig:recall}, Table~\ref{tab:t3} and Fig.~\ref{fig:main_thd_exp2}.

\textbf{Domain-specific Q$\&$A.} We first evaluate the win rates across different metrics for responses generated by \sysname against the four baseline RAG approaches for different relative metrics, as presented in Table \ref{tab:t1}. Note that the win rates represent the average results across all four edge nodes. The recall comparison with the ground-truths is recorded in Fig.~\ref{fig:recall}.

In the within-domain scenario, \sysname achieves a 59.3$\%$ overall win rate over Naive RAG, demonstrating that DGRAG’s KG organization is more effective at capturing complex relational structures within the local corpus, with particularly strong gains on the Comprehensiveness metric. Compared with Local RAG, which solely performs Local Query session of \sysname, our method shows an even larger advantage: by using the gate mechanism to assess the quality of local responses and dynamically decide whether to escalate the query, \sysname selectively leverages the more capable cloud LLM for a small portion of within-domain queries, achieving a 73.3$\%$ overall win rate while maintaining comparable reference-based recall. Since most within-domain queries are handled locally and \sysname employs an edge SLM, its performance does not surpass that of cloud RAG, which is entirely based on cross-edge retrieval and the cloud LLM. Nevertheless, across both evaluation metrics, DGRAG matches the answer quality of Cent. RAG that owns much heavier retrieval load, showing that its adaptive retrieval strategy in sufficient local knowledge does not compromise retrieval quality and even improves overall efficiency in distributed environments.

In the out-of-domain scenario, DGRAG’s advantage over both Naive RAG and Local RAG becomes even more pronounced, achieving win rates of 83.3$\%$ and 91.0$\%$ respectively. By accurately assessing the adequacy of local knowledge for a given query through the gate mechanism, DGRAG reliably escalates out-of-domain queries, performs cross-edge retrieval and eventually outputs more comprehensive response. Moreover, through the gate-controlled routing, DGRAG achieves performance comparable to the more powerful Cloud RAG while avoiding the full reliance on cloud resources. This dynamic decision-making further allows DGRAG to outperform Cent. RAG on out-of-domain queries with a 73.5$\%$ overall win rate, highlighting its effectiveness in selecting the appropriate retrieval scope and collaboration strategies. DGRAG performs noticeably better in the out-of-domain scenario compared to the within-domain scenario, which highlights the advantages and practical applicability of our approach in addressing the challenges of data decentralization and distributed retrieval, more closely aligned with the demands of real-world applications.

\begin{figure}[htp]
    \centering
    \includegraphics[width=0.85\columnwidth]{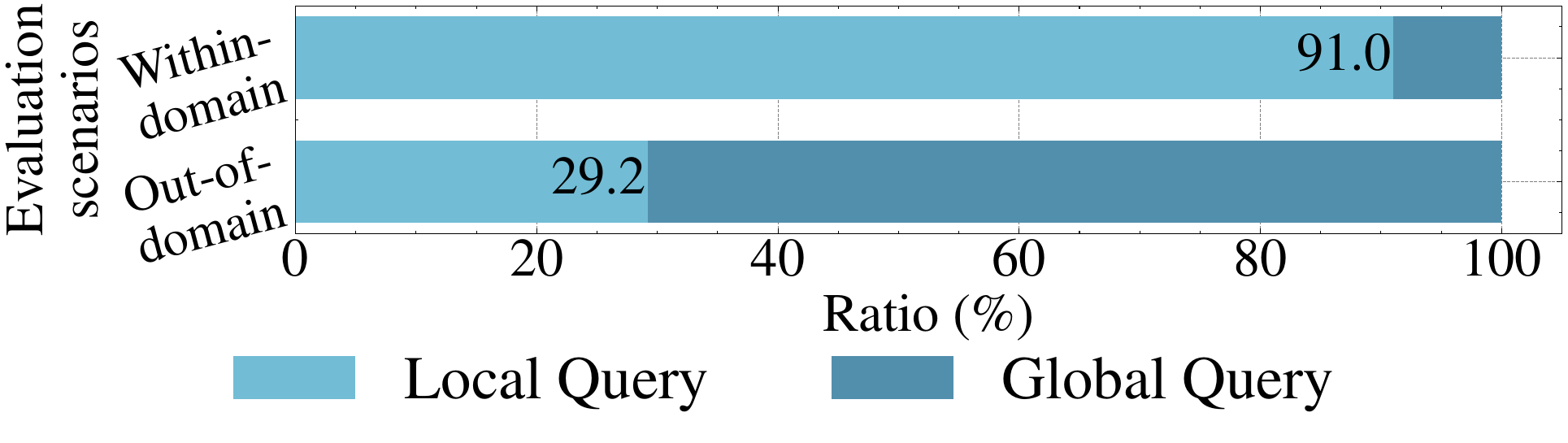}
    \caption{The ratio of queries operated locally or in the cloud across different evaluation scenarios in domain-specific Q$\&$A tasks.}
    \label{fig:ratio}
\end{figure}

Theoretically, within-domain queries should be resolved locally due to sufficient local knowledge and out-of-domain queries should be sent to the cloud server for cross-edge retrieval. However, as shown in Fig.~\ref{fig:ratio}, we observe that 9.0$\%$ of within-domain queries are nonetheless escalated to the cloud, indicating that the local SLMs still have performance limitations, which may fail to correctly interpret or utilize the local context and trigger the necessary cross-edge retrieval. While 29.2$\%$ of out-of-domain queries remain local, a consequence of overlapping cross-domain knowledge and the SLM’s capacity to generate plausible answers even without access to the most relevant knowledge.

\textbf{Mixed-domain Q$\&$A.} In mixed-domain Q$\&$A tasks, we similarly record the average win rates of DGRAG compared to other baseline RAG approaches in Table \ref{tab:t3} and the recall of different RAG approaches as metrics in Fig.~\ref{fig:main_thd_exp2} for effectiveness evaluation.

\begin{table}[t]
\centering
\caption{Average win rates of \sysname compared to baseline RAG approaches in mixed-domain Q$\&$A tasks.}
\label{tab:t3}
\resizebox{0.95\linewidth}{!}{
\begin{tabular}{lcccc} 
\toprule[1.5pt]
{\textbf{\begin{tabular}[l]{@{}l@{}}Method\\ Comparison\end{tabular}}} & \begin{tabular}[c]{@{}c@{}}DGRAG vs. \\Naïve RAG\end{tabular} & \begin{tabular}[c]{@{}c@{}}DGRAG vs.\\Local RAG\end{tabular} & \begin{tabular}[c]{@{}c@{}}DGRAG vs.\\Cloud RAG\end{tabular} & \begin{tabular}[c]{@{}c@{}}DGRAG vs.\\Cent. RAG\end{tabular}  \\ 
\midrule
Comprehensiveness & 82.5\% & 85.6\% & 66.3\% & 73.4\%\\
Diversity         & 71.4\% & 76.0\% & 44.8\% & 60.1\%\\
Empowerment       & 78.1\% & 82.9\% & 55.0\% & 66.6\%\\
\midrule
\textbf{Overall}  & \textbf{79.4\%} & \textbf{83.5\%} & \textbf{58.6\%} & \textbf{68.2\%} \\
\bottomrule[1.5pt]
\end{tabular}}
\end{table}

As shown in Table~\ref{tab:t3} and Fig.~\ref{fig:main_thd_exp2}, DGRAG outperforms both Naïve RAG and Local RAG across the two evaluation metrics. With domain knowledge distributed across edges, Naïve RAG often fails to retrieve the most relevant chunks from its limited local corpus, while Local RAG is even more susceptible to linking irrelevant KG elements or contexts, resulting in substantially degraded answer quality when generation is confined to the edge. In contrast, \sysname effectively exploits the complementary capabilities of edge nodes and the cloud via a gate mechanism that dynamically determines the most appropriate retrieval and generation source. When local knowledge is sufficient, \sysname relies on the best local response to maintain efficiency; when it is insufficient, \sysname escalates the query to the cloud, enabling cross-edge retrieval and cloud-based generation that provides more comprehensive and diverse responses, thereby ensuring answer quality. As the knowledge base expands, the introduction of thresholds helps control retrieval volume and thereby manage input context of the cloud LLM. This approach, combined with dynamic decision-making, yields a robust balance between efficiency and accuracy, resulting in consistently strong win rates and recall performance. Compared with stronger baselines, Cloud RAG benefits from full cloud-level cross-edge retrieval, yet DGRAG still attains a competitive win rate of 58.6$\%$ with only a marginal drop in recall, indicating that its local retrieval mechanism and the hierarchical query routing can approximate the performance of a purely cloud-based RAG system. While Cent. RAG achieves maximal coverage through complete knowledge centralization, DGRAG still secures a 68.2$\%$ win rate and comparable recall performance, showing that its dynamic decision-making and edge–cloud collaboration effectively compensate for the absence of full centralization. With precise retrieval enabled through both top-$m$ of the matched summaries and threshold controls, DGRAG can even surpass the performance of centralized RAG in distributed settings.

\subsection{Overhead}

\subsubsection{Time efficiency}

We record the average total time of the five RAG on both evaluation scenarios, as well as the time spent on key components of the \sysname pipeline, to comprehensively evaluate its time efficiency. Notably, the entire process involves four data transmissions: (1) Query Send: transmission of the query from the originating user to the server, (2) Request Send: transmission of the query from the server to other edge nodes, (3) Knowledge Send: transmission of the retrieved knowledge from the edges back to the server, and (4) Answer Send: transmission of the final response to the user. The network bandwidth is set to 50 Mbps via the router.

\begin{figure}[htp]
    \centering
    \includegraphics[width=1\columnwidth]{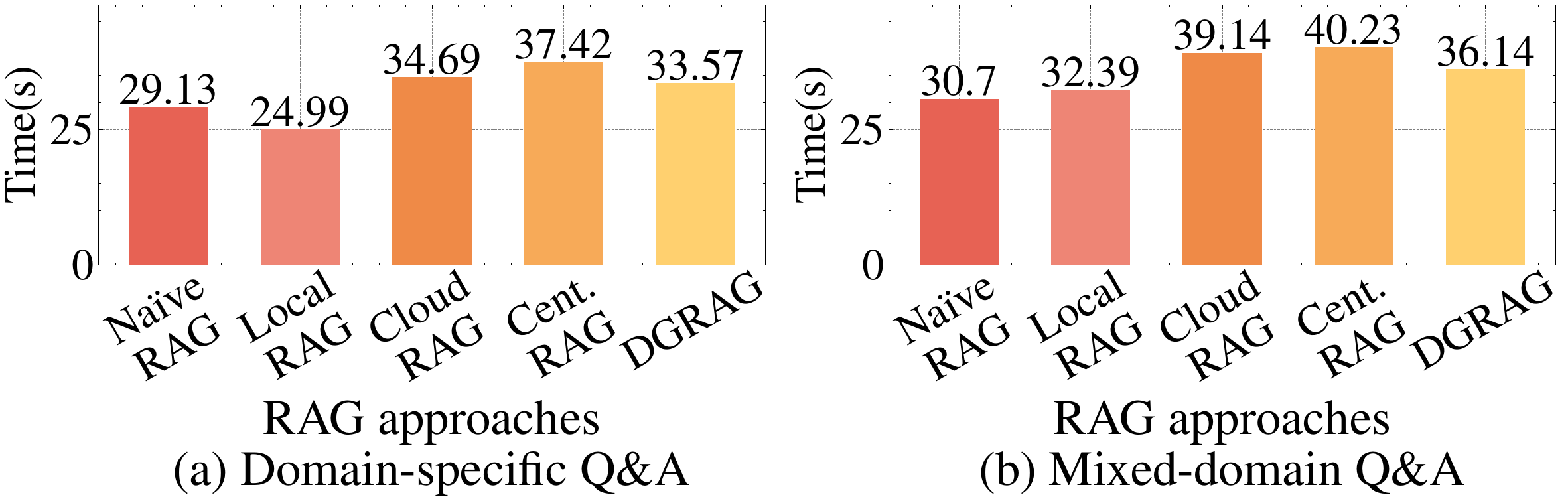}
    \caption{Average total time cost on (a) domain-specific Q$\&$A and (b) mixed-domain Q$\&$A scenarios by different RAG approaches.}
    \label{fig:time}
\end{figure}

\begin{figure}[htp]
    \centering
    \includegraphics[width=0.85\columnwidth]{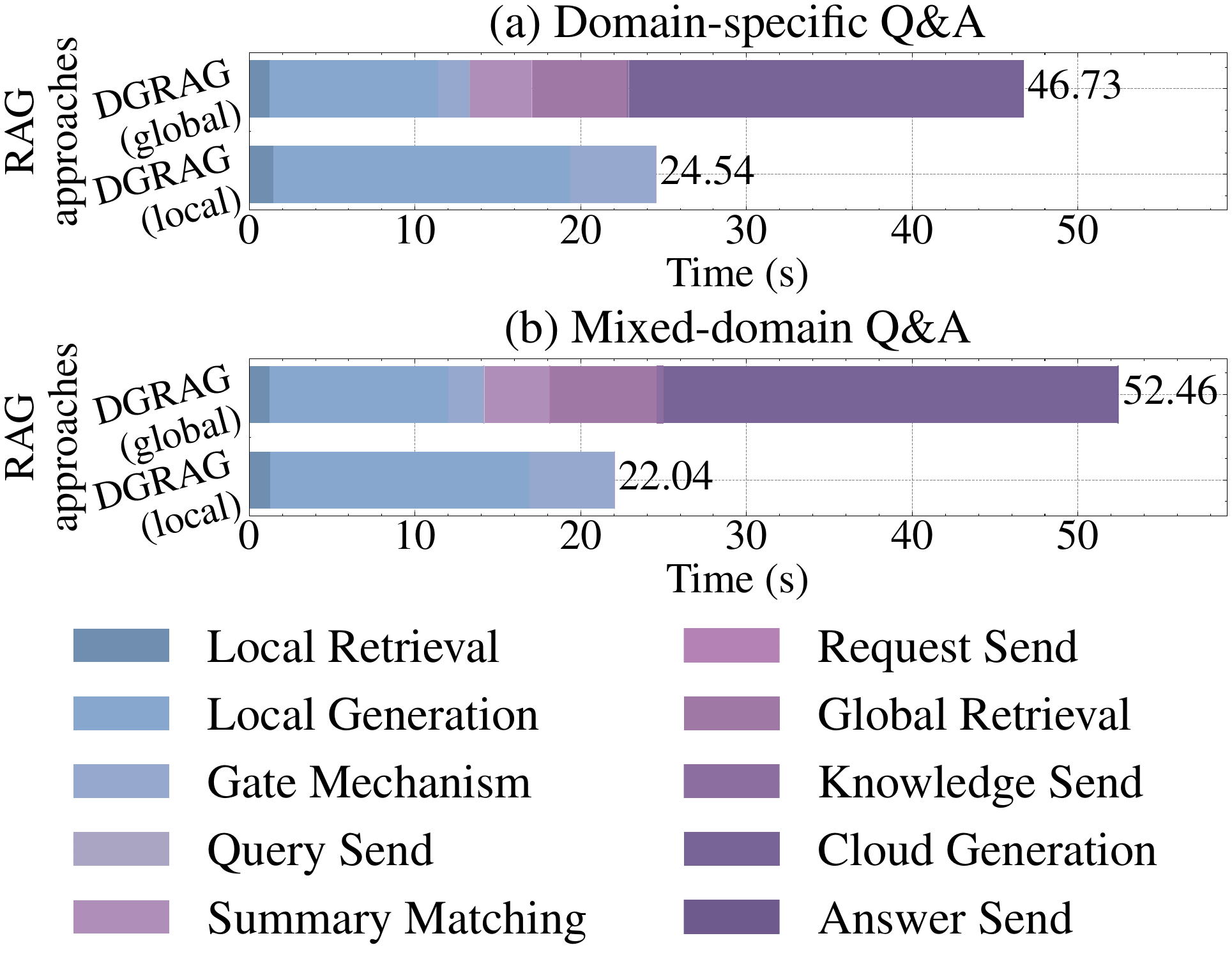}
    \caption{The time spent on key components of the \sysname pipeline in (a) domain-specific Q$\&$A and (b) mixed-domain Q$\&$A scenarios. Notably cross-edge retrieval mechanism consists of Summary Matching, Global Retrieval, Cloud Generation and several data transmission.}
    \label{fig:time_2}
\end{figure}

\textbf{Domain-specific Q$\&$A} Observed from Fig.~\ref{fig:time} (a), \sysname incurs a slightly higher average latency per query than Naïve RAG and Local RAG, yet remains noticeably faster than Cloud RAG and Cent. RAG. A closer component-level breakdown of the total time in Fig.~\ref{fig:time_2} (a) reveals the underlying reasons. In cases where local knowledge is sufficient (\textit{local}), the gate mechanism introduces some additional overhead because the edge SLM must select the best answer among multiple locally generated candidates. Nevertheless, aided by vLLM-based inference acceleration, batch inference in \sysname is actually faster than the single-response inference used in Naïve RAG and Local RAG. Overall, DGRAG(local) which includes local retrieval, generation, and judgement by the gate mechanism achieves lower end-to-end latency than both Naïve RAG and Local RAG, which also operate solely on the edge, and remains substantially faster than Cloud RAG, which requires additional cross-edge retrieval and edge-cloud cooperation, as well as Cent. RAG, which performs large-scale centralized retrieval. In the case of inadequate local knowledge (\textit{global}), the likelihood of generating an unsatisfactory local response increases, leading to a shorter Gate Mechanism time cost. In this scenario, Global Retrieval and Cloud Generation, key components of the cross-edge retrieval mechanism, account for a larger share of the overall time cost and consequently increase the total latency. However, due to the relatively small size of these transmitted data content, the communication latency is negligible. Overall, when local knowledge is insufficient, DGRAG increases additional latency due to edge–cloud collaboration. However, this overhead is well justified by the resulting performance gains and remains within an acceptable range for practical system efficiency.

\textbf{Mixed-domain Q$\&$A} Fig.~\ref{fig:time} and Fig.~\ref{fig:time_2} show that the time cost of mixed-domain Q$\&$A is broadly comparable to domain-specific Q$\&$A. The time spent in Local Query step increases due to the enlarged scale of the local knowledge base, while locally accelerated batch query not only reduces inference latency at the edge but also supports the decision of the gate mechanism. Cross-edge retrieval across multiple knowledge sources increases cloud consolidation and generation time, which is the primary factor behind DGRAG’s longer completion time in mixed-domain Q$\&$A compared with the domain-specific Q$\&$A evaluation. The threshold controls in Summary Matching and Global Retrieval sessions effectively limit the retrieval volume without compromising knowledge quality or relevance, leaving the communication cost minimal. Although the gate mechanism and the potential invocation of the cloud LLM inevitably make DGRAG slower than Naïve RAG and Local RAG, its adaptive routing of retrieval and generation allows DGRAG to deliver answer quality comparable to the more powerful Cloud RAG and Cent. RAG while still keeping the overall end-to-end latency well within a practical range.

\subsubsection{Resource consumption}

To assess the resource consumption of DGRAG compared to Cent. RAG, we focus on cloud storage requirements and LLM input tokens. First, we examine \textbf{cloud storage}. In the domain-specific Q$\&$A, DGRAG requires only 13.03 MB to store its summary database, whereas Cent. RAG, which integrates all edge knowledge bases into a centralized system, occupies 1.53 GB of memory. Similarly, DGRAG’s cloud server requires only 40.24 MB to store the embeddings of edge summaries, while centralizing the entire distributed corpus would demand 4.18 GB in mixed-domain Q$\&$A. This substantial gap arises from DGRAG’s design choice of sharing \emph{subgraph summaries} instead of raw documents: by organizing local knowledge into knowledge graphs and uploading only compact subgraph summaries, \sysname achieves efficient and economical knowledge sharing with minimal storage pressure on the cloud, while also avoiding direct exposure of full edge-side data. These lightweight summaries still preserve the key structural and semantic relationships required for effective retrieval, enabling the cloud to accurately identify relevant edges without maintaining a full replica of all edge knowledge.

Next, we calculate LLM input tokens. For the domain-specific Q$\&$A task, DGRAG reduces the total number of tokens processed by the cloud LLM from 178.6 million in Cent. RAG to 69.9 million—just 39$\%$ of the tokens required by Cent. RAG. In mixed-domain Q$\&$A tasks, DGRAG reduces the total tokens processed by the cloud LLM from 104.9 million to 37.9 million. In DGRAG’s hierarchical retrieval pipeline, queries are first handled locally, and only those deemed insufficiently supported by local knowledge are escalated to the cloud for cross-edge retrieval and generation. As shown in the \textit{overall performance}, this design still yields competitive win rates and recall compared to Cent. RAG, indicating that the reduced dependence on the cloud LLM does not come at the cost of answer quality. Moreover, the \textit{time efficiency} results demonstrate that \sysname completes end-to-end tasks with an average latency faster than Cent. RAG, confirming that the dynamic routing of the gate mechanism not only cuts computational and monetary costs at the cloud, but also maintains a favorable latency–quality trade-off in distributed edge–cloud environments.

\subsection{Ablation study}

\subsubsection{Subgraph size}
DGRAG utilizes subgraph summaries to distill edge information, and the quality of these summaries is closely tied to the subgraph size. To explore the impact of subgraph size on Summary Matching performance, we measure the subgraph size by the number of entities a single subgraph contains and examine how this parameter affects the Summary Matching session under the domain-specific Q$\&$A experimental setup.

\begin{figure}[htp]
    \centering
    \includegraphics[width=1\columnwidth]{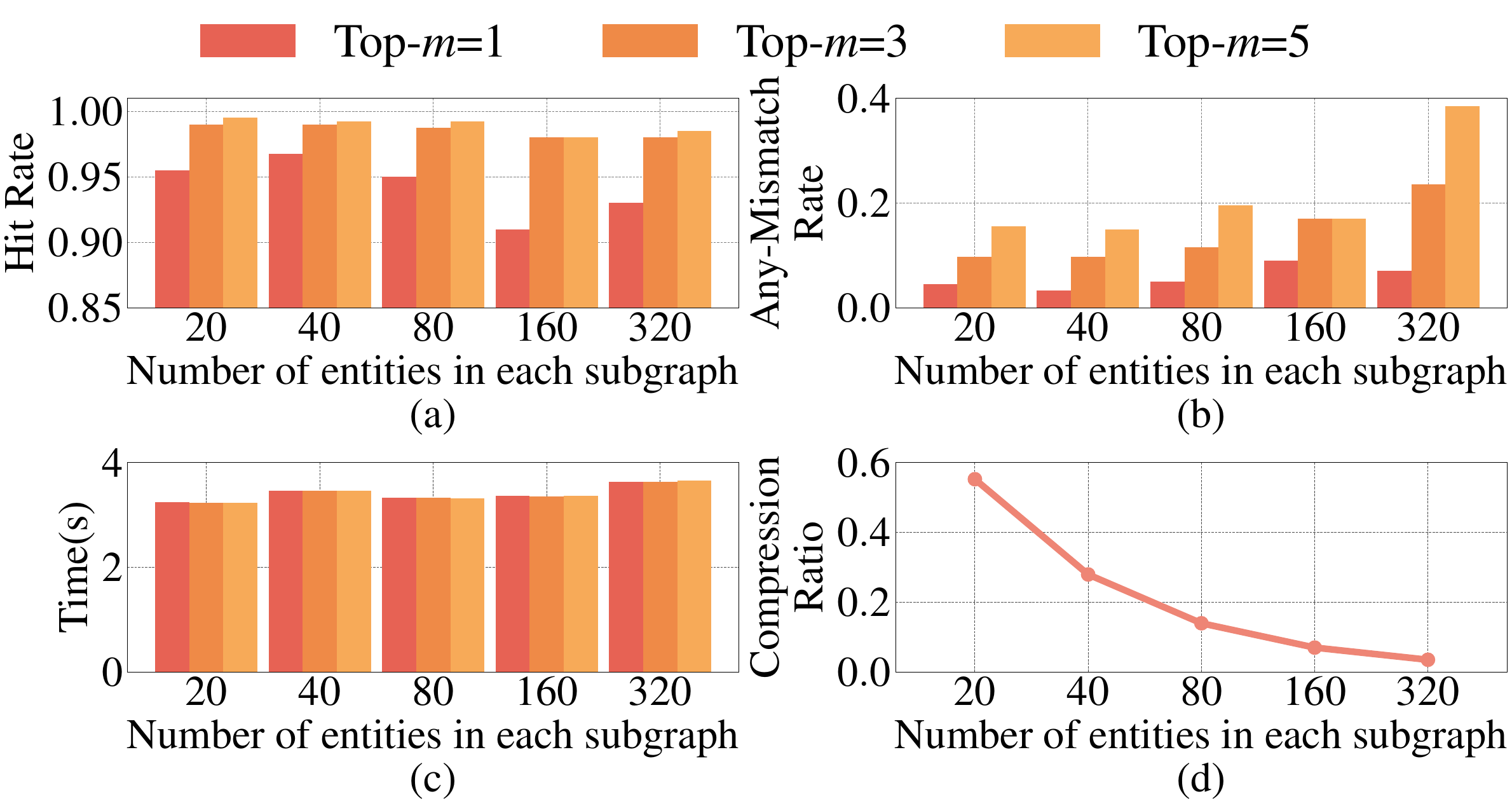}
    \caption{Effect of subgraph size on Summary Matching in different top-$m$ subgraph summary matching: (a) ``hit'' rate; (b) ``mismatch'' rate; (c) average query time; (d) the compression ratio of the edge storage between vector embeddings of subgraph summaries and raw textual knowledge embeddings.}
    \label{fig:hit rate}
\end{figure}

The 400 previously selected queries are matched against the summary database in the cloud based on the cosine similarity of their vector embeddings. A query is considered a ``hit'' if at least one of the top-$m$ retrieved subgraph summaries belongs to the same domain as the query and is considered a ``mismatch'' if the top-$m$ retrieved subgraph summaries exist non-intentional edge summary. As shown in Fig.~\ref{fig:hit rate}, under the condition of similar retrieval time, we set 40 as the number of entities in a single subgraph with its higher hit rate, lower any-mismatch rate and memory compression ratio.

\subsubsection{Thresholds}

To enhance DGRAG's efficiency in larger-scale retrieval, similarity thresholds are introduced in the \textbf{Summary Matching} and \textbf{Global Retrieval} sessions to control the retrieval expenses. The optimal threshold value is explored through experiments in mixed-domain Q$\&$A.

\begin{figure}[htp]
    \centering
    \includegraphics[width=1\columnwidth]{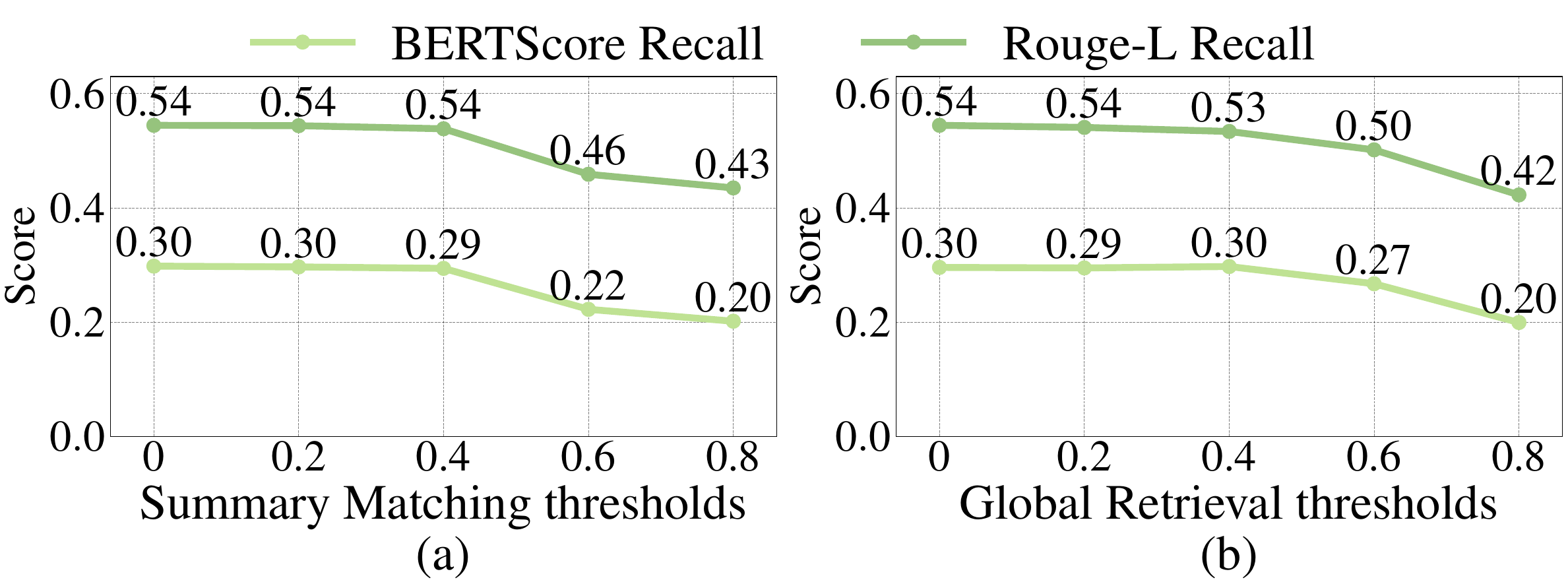}
    \caption{BERTScore Recall and Rouge-L Recall of responses under different (a) Summary Matching thresholds and (b) Global Retrieval thresholds in mixed-domain Q$\&$A tasks.}
    \label{fig:thd_exp2}
\end{figure}

We use the recall of answers as the evaluation metrics for the threshold selection. For the Summary Matching session, we first set the Global Retrieval threshold to 0, which represents returning all relevant knowledge from retrieval edges to the cloud server. Then, in the given range of top-$m$ candidates, only edges belonging to summaries that exceed the Summary Matching threshold are selected for the Global Retrieval session. Fig.~\ref{fig:thd_exp2} shows that both the BERTScore Recall and Rouge-L Recall begin to drop when the threshold surpasses 0.4. Therefore, we set the Summary Matching threshold as 0.4 to reduce retrieval expenses and filter knowledge sources efficiently. With the Summary Matching threshold frozen, we perform the same experiment for the Global Retrieval threshold, i.e., only the knowledge that exceeds the Global Retrieval threshold is returned and the contexts are re-ranked by descending cosine similarity. According to the experimental results in Fig.~\ref{fig:thd_exp2}, the recall value gradually decreases when the threshold exceeds 0.4. In order to efficiently integrate and utilize the edge knowledge, the Global Retrieval threshold is adopted to 0.4, which curbs data transmission, reduces latency, and incidentally strengthens privacy by limiting the volume of edge knowledge exposed to the cloud.

\subsubsection{Key components}

We also conduct an ablation study to assess the contribution of key components of DGRAG to its overall performance. The domain-specific Q$\&$A experimental setup is chosen for this analysis, as it offers a more straightforward and focused demonstration of DGRAG's workflow, and we select the average win rate as the metric. The results of this study are presented in Table \ref{tab:ablation}. Notably \textit{w/o CD} and \textit{w/o SE} denote \sysname without the confidence detection and similarity evaluation components of the gate mechanism, respectively. \textit{w/o BQ} represents \sysname using only single inference instead of batch inference; in this case, the gate mechanism makes decisions based solely on confidence detection. \textit{w/o KG} indicates that retrieval in both Local Query and the Cross-edge Retrieval Mechanism relies exclusively on the vector database, without using the graph database, while keeping the value of $k$ in top-$k$ retrieval unchanged.
\begin{table*}[ht]
\centering
\caption{Average win rates of \sysname compared to baseline RAG approaches in the ablation study of key components.}
\label{tab:ablation}
\begin{tabular}{lcccccccc}
\toprule[1.5pt]
\multicolumn{1}{l}{\textbf{Scenario}}                                                    & \multicolumn{4}{c}{\textbf{Within-domain}}                         & \multicolumn{4}{c}{\textbf{Out-of-domain}}\\ \midrule
\multicolumn{1}{l}{\textbf{\begin{tabular}[l]{@{}l@{}}Method\\ Comparison\end{tabular}}} & \begin{tabular}[c]{@{}c@{}}DGRAG vs. \\ w/o CD\end{tabular} & \begin{tabular}[c]{@{}c@{}}DGRAG vs.\\ w/o SE\end{tabular} & \begin{tabular}[c]{@{}c@{}}DGRAG vs.\\ w/o BQ\end{tabular} & \begin{tabular}[c]{@{}c@{}}DGRAG vs.\\ w/o KG\end{tabular} & \begin{tabular}[c]{@{}c@{}}DGRAG vs. \\ w/o CD\end{tabular} & \begin{tabular}[c]{@{}c@{}}DGRAG vs.\\ w/o SE\end{tabular} & \begin{tabular}[c]{@{}c@{}}DGRAG vs.\\ w/o BQ\end{tabular} & \begin{tabular}[c]{@{}c@{}}DGRAG vs.\\ w/o KG\end{tabular} \\ \midrule
Comprehensiveness & 82.8\%                                                    & 79.0\%                                                    & 80.5\%                                                    & 65.3\%                                                    & 86.0\%                                                    & 80.5\%                                                     & 77.5\%                                                     & 82.3\%                                                      \\
Diversity         & 57.8\%                                                    & 56.8\%                                                    & 59.3\%                                                    & 40.0\%                                                    & 80.3\%                                                    & 71.5\%                                                     & 71.3\%                                                     & 73.0\%                                                      \\
Empowerment       & 73.3\%                                                    & 71.0\%                                                    & 72.8\%                                                    & 52.3\%                                                    & 83.8\%                                                    & 77.8\%                                                     & 76.3\%                                                     & 78.5\%                                                      \\
\midrule
\textbf{Overall}                                                                         & \textbf{75.3\%}                                                    & \textbf{74.0\%}                                                    & \textbf{74.0\%}                                                    & \textbf{54.8\%}                                                    & \textbf{83.8\%}                                                    & \textbf{78.5\%}                                                     & \textbf{76.5\%}                                                     & \textbf{78.8\%}\\ 
\bottomrule[1.5pt]
\end{tabular}
\end{table*}

\textbf{Confidence Detection (CD) and Similarity Evaluation (SE) in Gate Mechanism.} Table \ref{tab:ablation} shows that removing either the confidence detection (CD) or similarity evaluation (SE) component significantly degrade the performance of \sysname, with the removal of the confidence detection phase causing a more pronounced decline. These two components serve complementary roles in the operation of the gate mechanism. Without the confidence detection phase, local responses with high similarity but low confidence are misjudged by the similarity evaluation phase, preventing the system from initiating further retrieval. Conversely, omitting the SE phase makes the gate mechanism difficult to detect hallucinated responses generated by the edge SLM based on certain but insufficient knowledge. This issue is especially prominent in out-of-domain queries, where knowledge gaps are more likely to occur.

\textbf{Batch inference in Local Query (BQ).} 
Removing the batch inference in Local Query (BQ) significantly affects the response quality for both within-domain and out-of-domain queries, as shown in Table \ref{tab:ablation}. Determining the adequacy of local knowledge based on a single query is inherently prone to sampling uncertainty, which can lead to misjudgments about whether the local knowledge is sufficient to generate a reliable response. In contrast, the use of batch inference during the Local Query step aggregates responses from multiple queries, providing a more robust and reliable basis for the subsequent gate mechanism to assess the adequacy of local knowledge. This preliminary yet essential component contributes significantly to DGRAG's overall performance, improving both the reliability and quality of the generated answers.

\textbf{Knowledge graph-based retrieval(KG).} When \sysname relies solely on dense-vector search, its performance drops significantly, as embeddings alone fail to capture the latent, higher-order relationships between entities. Specifically, in the within-domain scenario, removing the knowledge graph-based retrieval and using only vector search has a limited impact, primarily affecting the comprehensiveness metric. This is because within-domain queries typically benefit from sufficient local knowledge, where traditional vector retrieval can handle a portion of the domain-specific queries effectively. However, in the out-of-domain scenario, vector search struggles with the complexity of queries from unfamiliar domains, which are then handled by the cross-edge retrieval mechanism. Even when performing the cross-edge retrieval pipeline in highly related edge nodes, vector search still underperforms compared to knowledge graph-based methods, leading to a marked difference in performance. This degradation underscores the indispensability of the knowledge graph component, whose explicit relational structure captures abstract associations and enables the system to retrieve the interconnected evidence necessary for accurately answering complex queries.

%% file: sections/Conclusion.tex
\section{Conclusion}
This paper introduces DGRAG, a distributed, graph-based RAG framework that addresses the limitations of centralized retrieval in privacy-sensitive, resource-constrained edge–cloud environments. DGRAG organizes local knowledge at each edge into knowledge graphs and shares only subgraph summaries with the cloud, enabling lightweight and privacy-preserving global indexing. During inference, a gate mechanism evaluates local response confidence and similarity to decide whether a query can be answered locally or should be escalated. For escalated queries, a cross-edge retrieval mechanism identifies relevant edges via summary matching and leverages cloud-level generation to produce accurate and comprehensive answers.

Through this workflow, DGRAG achieves an effective balance between local autonomy and global collaboration, reducing storage and token costs while maintaining strong answer quality. Experimental results demonstrate that DGRAG consistently outperforms decentralized baselines and achieves performance comparable to centralized systems with significantly lower resource consumption. These findings highlight DGRAG as a practical and scalable foundation for distributed RAG in real-world, heterogeneous knowledge environments.